\documentclass[lettersize,journal]{IEEEtran}
\usepackage{amsmath,amsfonts}
\usepackage{algorithmic}
\usepackage{algorithm}
\usepackage{array}
\usepackage{amssymb}
\usepackage{multirow}
\usepackage{multicol}
\usepackage{booktabs}
\usepackage{colortbl}   
\usepackage{xcolor}
\usepackage[caption=false,font=normalsize,labelfont=sf,textfont=sf]{subfig}
\usepackage{textcomp}
\usepackage{stfloats}
\usepackage{url}
\usepackage{verbatim}
\usepackage{graphicx}
\usepackage{cite}
\begin{document}

\title{Enhanced Knowledge Distillation for Detection Transformers via Teacher Prediction Refinement}

\author{Yitong Xing, Yuhao Cheng, Yanping Li, Yichao Yan
\thanks{Yitong Xing, Yuhao Cheng, Yanping Li, and Yichao Yan are with the School of Computer Science, Shanghai Jiao Tong University, Shanghai 200240, China (e-mail: richardxyt@sjtu.edu.cn, chengyuhao@sjtu.edu.cn, ypli2024@sjtu.edu.cn, yanyichao@sjtu.edu.cn).}
\thanks{Corresponding author: Yichao Yan.}
}

\markboth{Journal of \LaTeX\ Class Files,~Vol.~14, No.~8, August~2021}%
{Shell \MakeLowercase{\textit{et al.}}: A Sample Article Using I
。EEEtran.cls for IEEE Journals}


\maketitle

\begin{abstract}
Detection Transformers (DETRs) achieve strong performance in object detection but remain challenging to deploy on edge devices due to their high computational cost. Existing DETR distillation methods mainly focus on aligning distillation points, while largely overlooking the quality of the teacher’s supervision itself. We observe that due to stage-wise non-monotonic prediction behavior in DETRs, well-localized or correctly classified predictions from earlier stages may degrade in later ones, and some negative predictions become increasingly overconfident. As a result, relying solely on the current stage’s predictions yields inaccurate and inconsistent supervision. To address this issue, we propose Teacher Prediction Refinement Distillation (TPRD), a plug-and-play module that refines teacher predictions before distillation by exploiting stage-wise prediction information. TPRD improves supervision quality through Positive Prediction Correction (PPC), which corrects degraded positive predictions by restoring more accurate ones from earlier stages, ensuring reliable localization and classification signals, and Negative Prediction Suppression (NPS) suppresses the influence of overconfident negatives, preventing them from providing misleading supervision to the student. To preserve informative dark knowledge, we further introduce Maximum Dark Knowledge Preservation (MDKP), which selectively refines target-class logits while retaining non-target relations. Extensive experiments on MS COCO and Pascal VOC demonstrate the effectiveness and robustness of the proposed method. Our code is available at \url{https://github.com/xingyitong1/TPRD}. 
\end{abstract}


\section{Introduction}
\IEEEPARstart{O}{bject} detection is one of the most critical tasks in computer vision. Over time, detector performance has improved remarkably, especially with the emergence of DETRs [1]–[5], which employ transformer-based architectures for effective detection. However, these models have been accompanied by unaffordable parameters and computational cost, making direct deployment on edge devices challenging. Knowledge distillation\cite{hinton2015distilling}
has emerged as an effective approach for model compression and performance enhancement. Consequently, knowledge distillation for DETR models has attracted increasing research attention. 
\begin{figure}
    \centering
    \includegraphics[width=0.5\textwidth]{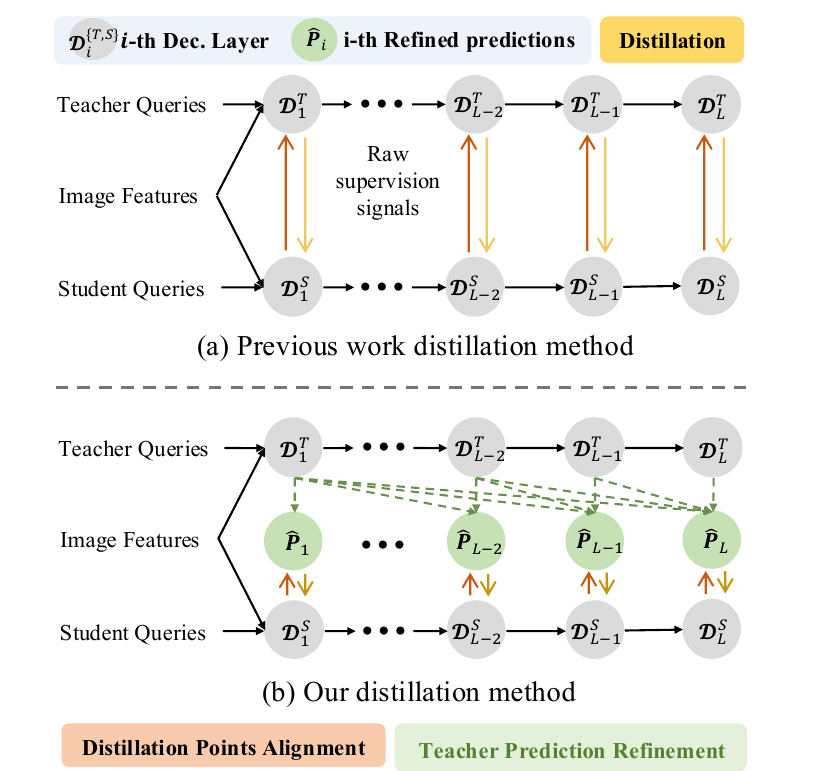}
    \caption{Comparison between previous methods and our method. (a) Previous methods directly use the raw outputs from each stage as supervision signals. (b) In contrast, our method refines teacher predictions for each stage by leveraging information from all preceding stages to provide more reliable supervision.}
    \label{fig:fig1}
\end{figure}

Unlike CNN-based detectors, DETRs employ an unordered set of object queries for prediction, resulting in no direct one-to-one correspondence between teacher and student predictions. This misalignment introduces unique challenges for DETR distillation. To address this challenge, several studies \cite{chang2023detrdistill,ChenD3ETR,liu2024knowledge} adopt the Hungarian matching algorithm to adaptively align teacher and student predictions, while \cite{wang2024kd} introduces a shared set of distillation queries to achieve a natural one-to-one alignment. Although these methods effectively mitigate prediction misalignment, they largely assume that the teacher’s predictions used for distillation are inherently reliable, and directly employ the raw outputs of the current decoder stage as supervision.

\begin{figure*}[t]
  \centering
  \includegraphics[width=\textwidth]{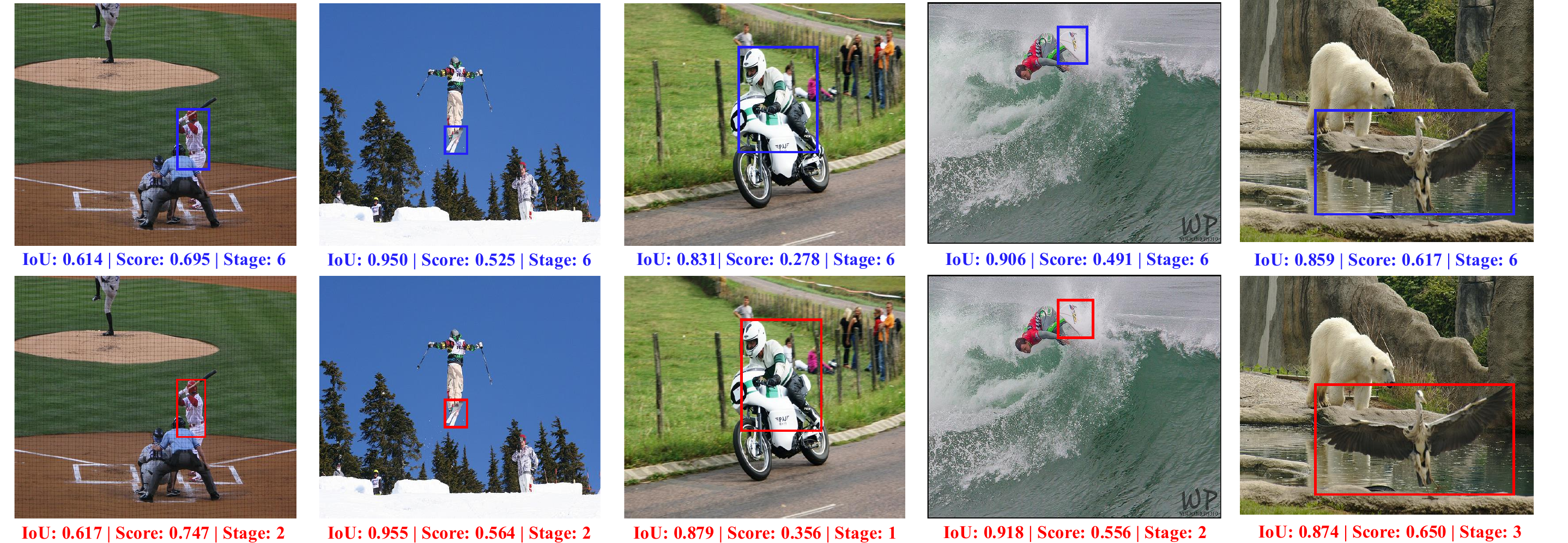}
  \caption{Visualization of the non-monotonic prediction behavior during DETR training. We illustrate the Hungarian matching results across different decoder stages to investigate whether earlier-stage predictions can surpass later ones. In several cases, earlier predictions exhibit higher IoU or classification confidence than the final-stage outputs, confirming that prediction quality does not monotonically improve across stages.}
  \label{fig:fig2}
\end{figure*}

However, this assumption does not always hold in DETRs. Due to the cascading update mechanism of the decoder, prediction quality does not necessarily improve monotonically across stages. Previous work~\cite{chen2023enhanced} has shown that predictions from intermediate stages can be superior to final-stage predictions and has exploited this observation to improve query-based detector training. However, its focus is on query recollection for detector optimization, while the influence of stage-wise prediction variation on teacher supervision quality in DETR distillation has not been sufficiently investigated. To investigate this phenomenon, we visualize the Hungarian matching results across multiple stages for several samples. As shown in Figure \ref{fig:fig2}, earlier stage predictions can indeed outperform those from the final stage. Furthermore, to verify that this behavior is neither incidental nor restricted to the last stage, we perform additional analyses in Section \ref{Section:Methods}. These consistent findings demonstrate that such behavior also persists across all stages rather than the final layer. Consequently, directly using the raw supervision signals from the teacher’s current stage, as done in previous works, leads to a suboptimal situation where more accurate signals from earlier stages are neglected, limiting the full leverage of the teacher model’s knowledge.

These observations suggest that effective DETR distillation should not only focus on aligning distillation points but also on refining which predictions are distilled, especially under stage-wise uncertainty. Motivated by this, we propose the \textbf{Teacher Prediction Refinement Distillation (TPRD)} mechanism, whose primary goal is to refine the supervision signals provided by the teacher model by identifying more reliable predictions across stages. During training, predictions are divided into positives and negatives via Hungarian matching. \textbf{Positive Prediction Correction (PPC)}: Since positive predictions with higher IoU and confidence scores generally provide more reliable supervision, we introduce a correction step to refine degraded positives. Specifically, when a positive prediction at the current stage exhibits degraded localization or classification quality, it is \textit{replaced} with its counterpart from an earlier stage that achieves higher IoU and confidence, thereby restoring more reliable supervision. \textbf{Negative Prediction Suppression (NPS)}: Negative predictions with high confidence scores may introduce misleading supervision signals that contradict the ground truth and hinder the student’s learning. To alleviate this issue, we suppress such noisy supervision by \textit{replacing} overconfident negative predictions with those from earlier stages that exhibit lower confidence scores. Consequently, employing these refined teacher predictions as supervision signals leads to more stable and effective distillation.

To minimize the potential side effects introduced by the aforementioned \textit{replacement} operation, we adopt the \textbf{Maximum Dark Knowledge Preservation (MDKP)} strategy. Specifically, we only replace the logits of the target class, leaving those of the others unchanged. In addition, to preserve the inter-class relationships among non-target classes, we employ the Decoupled KL divergence \cite{zhao2022decoupled}. This strategy allows TPRD to refine supervision signals while retaining the teacher model’s dark knowledge.

In summary, the contributions of this paper are as follows:
\begin{itemize}
    \item We extend the study of stage-wise non-monotonic prediction behavior to DETR distillation and show that fluctuations across decoder stages can result in unreliable teacher supervision.
    \item We propose Teacher Prediction Refinement Distillation, which refines teacher supervision by correcting degraded positive predictions and suppressing overconfident negatives while preserving dark knowledge.
    \item Extensive experiments on MS COCO and Pascal VOC demonstrate that TPRD consistently improves multiple DETR distillation frameworks across different DETR variants, validating its effectiveness and robustness.
\end{itemize}

\section{Related Works}
\subsection{DETR families}
DETR~\cite{carion2020end} is the first end-to-end transformer-based object detector, formulating detection as a set prediction problem with learnable queries and bipartite matching, thereby removing hand-crafted components such as anchors and NMS. Despite its conceptual simplicity, DETR suffers from slow convergence, high computational cost, and training instability, motivating a series of follow-up works~\cite{zhu2020deformable,meng2021conditional,wang2022anchor,liu2022dab,li2022dn,zhang2022dino,chen2023group,jia2023detrs,hou2024relation}. To accelerate convergence and stabilize training, Deformable DETR~\cite{zhu2020deformable} introduced deformable attention and multi-scale feature aggregation, while Anchor DETR~\cite{wang2022anchor} and DAB-DETR~\cite{liu2022dab} injected spatial priors into query design. DN-DETR~\cite{li2022dn} and DINO~\cite{zhang2022dino} further improved optimization stability via query denoising, whereas Group-DETR~\cite{chen2023group} and H-DETR~\cite{jia2023detrs} alleviated sparse supervision by adopting one-to-many matching. More recently, Relation-DETR~\cite{hou2024relation} enhanced convergence and spatial reasoning by incorporating positional relation priors.  

Another line of research focuses on reducing the computational overhead of DETRs. Sparse-DETR~\cite{roh2021sparse} pruned redundant tokens to lower encoder complexity, and Lite-DETR~\cite{li2023lite} improved efficiency through interleaved feature encoding. Recent lightweight variants, including RT-DETR~\cite{zhao2024detrs}, D-FINE~\cite{peng2024d}, and DEIM~\cite{huang2025deim}, have achieved real-time performance and even surpassed the YOLO series~\cite{tian2025yolov12attentioncentricrealtimeobject}, highlighting the practical maturity of DETR-based detectors.  

Beyond architectural optimization, knowledge distillation has emerged as an effective complementary strategy for deploying lightweight DETRs. Our work builds on this direction by improving distillation effectiveness through refining the quality of teacher supervision.

\subsection{Knowledge distillation for CNN-based detectors}
Knowledge distillation (KD) was first introduced by Hinton et al.~\cite{hinton2015distilling}, who minimized the KL divergence between teacher and student logits. Subsequent studies mainly evolved along two directions: feature-based distillation~\cite{romero2015fitnetshintsdeepnets,zagoruyko2016paying,tian2019contrastive,chen2021distilling,liu2023function,guo2023class,park2019relational,song2022spot}, which transfers intermediate representations, and logit-based distillation~\cite{zhao2022decoupled,chi2023normkd,jin2023multi,xu2025local,zhang2024cross,sun2024logit,li2021reskd,zhu2025ckd}, which aligns output distributions. These techniques have been widely extended from classification to dense prediction tasks, including object detection. Early attempts at detection distillation focused on transferring neck and head features~\cite{chen2017learning}. To address foreground–background imbalance and uneven pixel contributions, subsequent works introduced various adaptive masking and weighting strategies during feature distillation~\cite{guo2021distilling,zhang2023structured,yang2022focal,huang2022masked,yang2022masked,lan2025acam,wang2024relation}. Representative methods include DeFeat~\cite{guo2021distilling}, which decouples foreground and background distillation, FKD~\cite{zhang2023structured}, which employs spatial and channel attention masks, and MasKD~\cite{huang2022masked} and ACAM-KD~\cite{lan2025acam}, which further introduce learnable or conditional masking mechanisms.  

In parallel, several studies explored distillation at the detection head level. LD~\cite{zheng2022localization} emphasized localization-aware knowledge transfer by focusing on high-confidence regions, while CrossKD~\cite{wang2023crosskd} mitigated gradient conflicts by sharing the teacher’s detection head during logit distillation.  

Despite their effectiveness, all the above methods were developed for CNN-based detectors and rely on dense feature or anchor-based supervision. Such designs are fundamentally incompatible with DETR architectures, which perform object detection via sparse object queries and set-based bipartite matching.

\subsection{Knowledge distillation for DETRs}
The main difference between DETR distillation and CNN-based methods lies in the issue of consistent distillation points \cite{wang2024kd}. Due to the sliding window-based feature extraction mechanism of CNNs and their direct predictions on feature maps, there exists a straightforward spatial correspondence between teacher and student predictions. In contrast, DETR employs a set of unordered queries to detect objects, resulting in no one-to-one relationship between teacher and student predictions. Therefore, most CNN-based distillation methods cannot be directly applied to DETR.

To address this issue, DETRDistill\cite{chang2023detrdistill} reused bipartite matching between teacher and student queries to ensure consistency in distillation points. $\text{D}^3$ETR\cite{ChenD3ETR} mitigated the instability of bipartite matching by introducing an additional fixed matching strategy. KD-DETR\cite{wang2024kd} observed that using bipartite matching still suffered from training instability and computational complexity, and thus proposed employing shared teacher–student object queries as consistent distillation points to avoid bipartite matching. QSKD\cite{liu2024knowledge} further analyzed the roles of different queries and selected only useful ones (positive and hard negative) for distillation, significantly reducing computational overhead. In addition, CLoCKDistill\cite{lan2025clockdistill} enhanced distillation consistency by incorporating location- and context-aware query alignment, while OD-DETR\cite{wu2024od} stabilized DETR training through an online distillation scheme that dynamically updates the teacher during training.

Although the above methods effectively addressed the issue of consistent distillation points, the knowledge, or more precisely, the supervision signals they utilized, remained relatively coarse, and potentially more reliable supervision signals available in the teacher’s intermediate predictions are not fully utilized. To overcome this limitation, we propose a Teacher Prediction Refinement Distillation approach that improves supervision quality by selectively refining teacher predictions across stages before distillation.

\begin{figure*}[t]
  \centering
  \includegraphics[width=\textwidth]{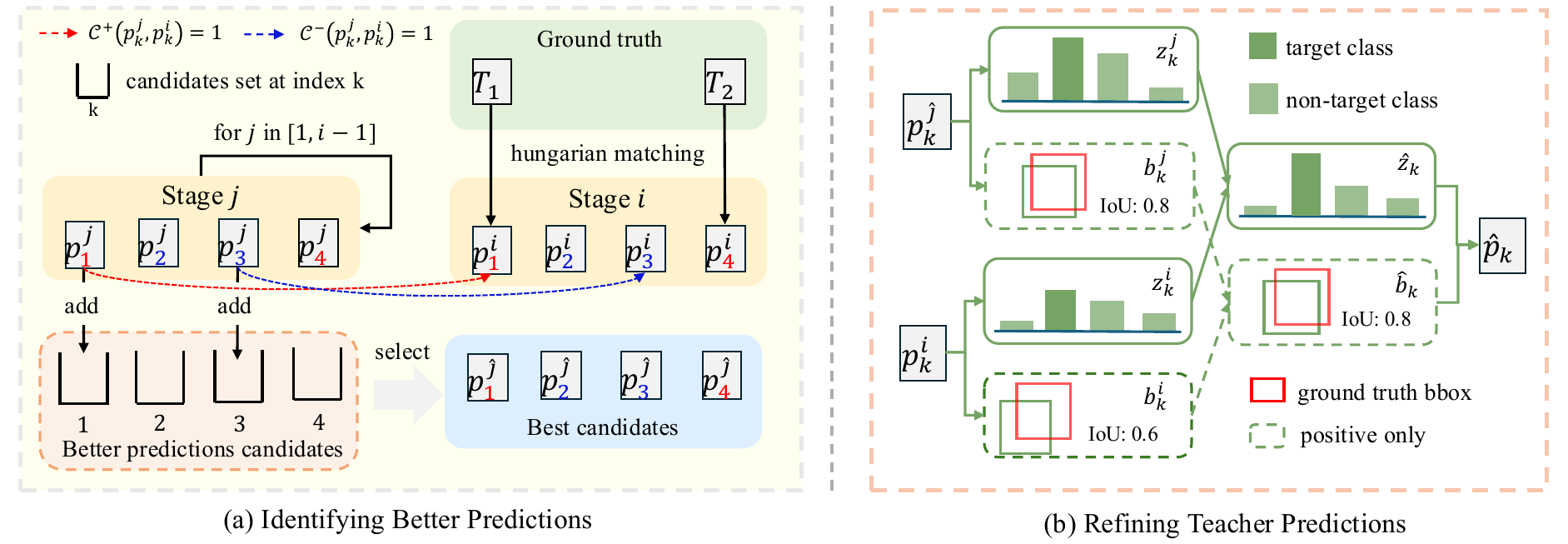}
  \caption{Overview of our proposed method. (a) Identifying better predictions: Each previous stage $j$ compares its predictions with those of the current stage $i$ based on Eq. (\ref{eq:6}) and Eq. (\ref{eq:9}). Satisfied predictions are added to candidate sets, from which the best candidates are selected to refine the current predictions. (b) Refining teacher predictions: We replace the target-class logits of current predictions with those from the best candidates, keeping other logits unchanged. For positive predictions, bounding boxes are also replaced.}
  \label{fig:fig3}
\end{figure*}

\section{Methods}
\label{Section:Methods}
In this section, we first introduce the preliminaries, including the cascading update mechanism of DETRs and the definitions of positive and negative predictions. We then present our motivation through empirical analysis. Based on this motivation, we describe our proposed TPRD framework, which refines teacher supervision through two steps: identifying better predictions and refining teacher predictions accordingly. Finally, we present the overall training objective of our method.
\subsection{Preliminaries}
\subsubsection{Cascading update mechanism in DETRs}
DETRs typically consist of three components: Backbone $\mathcal{B}$, Encoder $\mathcal{E}$, and Decoder $\mathcal{D}$. Given an input image $\mathbf{x} \in \mathbb{R}^{H \times W \times 3}$, the backbone and encoder extract image features $\mathbf{F} \in \mathbb{R}^{HW \times d}$, which are then fed into the decoder $\mathcal{D}$ with a set of learnable object queries $\{\mathbf{q}_i\}_{i=1}^N$, where $N$ denotes the number of queries and $d$ is the hidden dimension. Decoder $\mathcal{D}$ consists of a series of stages $\{D^i\}_{i=1}^L$ ($L$ is typically six), each sequentially containing: 1) a self-attention layer to model global interactions among queries; 2) a cross-attention layer to aggregate image features for each query; 3) a feed-forward network (FFN) followed by a prediction head that decodes queries into class probabilities and predicted bounding boxes $\{\mathbf{p}_i = (\mathbf{c}_i,\mathbf{b}_i)\}_{i=1}^N$. 

In the decoder, queries are processed in a cascaded manner, with each stage refining the output from the previous one:
\begin{equation}
    \label{Eq:1}
    \mathbf{q}_i^j = D^j(\mathbf{q}_i^{j-1},\mathbf{F}) + \mathbf{q}_i^{j-1},
\end{equation}
where $j$ is the stage index. In addition, many DETR variants further employ the box refinement strategy, where each stage predicts a residual offset relative to the previous bounding box:
\begin{equation}
\label{eq:2}
\mathbf{b}_i^j = \operatorname{MLP}_{\text{box}}(\mathbf{q}_i^j) + \mathbf{b}_i^{j-1},
\end{equation}
where $\operatorname{MLP}_{\text{box}}$ denotes the box prediction head. This cascading update mechanism implies that each stage’s outputs depend only on the preceding stage's outputs, causing both positive and negative updates to propagate to subsequent stages.

\subsubsection{Positive and Negative Predictions} 
During training, DETRs employ the Hungarian matching algorithm to find the optimal one-to-one correspondence between GTs $ \{y_i=(\mathbf{c}_i^{gt},\mathbf{b}_i^{gt})\}_{i=1}^{N}$ (padded with $\varnothing$) and the predictions $\{\mathbf{p}_i \}_{i=1}^N$. The optimization process can be represented as follows:
\begin{equation}
\begin{aligned}
    \label{Eq:3}
    \hat{\sigma} = \mathop{\arg\min}_{\sigma} \space & \sum_i^N [\lambda_{cls}^{cost} \mathcal{L}_{cls}^{cost}(\mathbf{c}_i,\mathbf{c}^{gt}_{\sigma(i)}) \\
    & + \mathbb{I}(\sigma(i) \neq \varnothing)\lambda_{box}^{cost}\mathcal{L}_{box}^{cost} (\mathbf{b}_i,\mathbf{b}^{gt}_{\sigma(i)})],
\end{aligned}
\end{equation}
where $\sigma$ denotes the permutation of the GTs, $\hat{\sigma}$ is the optimal permutation. $\mathcal{L}_{cls}^{cost} \text{ and } \mathcal{L}_{box}^{cost}$ are typically focal loss and the combination of L1 loss and GIoU loss.

After Hungarian matching, the predictions are divided into two groups: predictions assigned to a ground-truth object are defined as positive, and the remaining are negative:
\begin{equation}
    \label{eq:4}
    \mathbf{P}^{+} = \{\mathbf{p}_i | \space \hat{\sigma}(i) \neq \varnothing\}, \mathbf{P}^{-} = \{\mathbf{p}_i | \space \hat{\sigma}(i) = \varnothing \}.
\end{equation}
Correspondingly, the indices for these predictions can be denoted as:
\begin{equation}
    \label{eq:5}
    \mathbf{I}^{+} = \{i | \space \hat{\sigma}(i) \neq \varnothing\},\mathbf{I}^{-} = \{i | \space \hat{\sigma}(i) = \varnothing\}.
\end{equation}
This definition serves as the foundation for the subsequent supervision signals refinement in the distillation process. 

\begin{table}[t]
\footnotesize
    \centering
    \caption{Error propagation rates of different stages on different DETR variants. ``Pos." denotes positive predictions, ``Neg." denotes negative predictions.}
    \label{tab:tab1}
    \resizebox{0.5\textwidth}{!}{
        \begin{tabular}{c|cc|cc|cc}
            \toprule
            \multirow{2}{*}{Stage} & 
            \multicolumn{2}{c|}{Deformable DETR} & 
            \multicolumn{2}{c|}{DAB-DETR} & 
            \multicolumn{2}{c}{DINO} \\
            & Pos. & Neg. & Pos. & Neg. & Pos. & Neg. \\
            \midrule
            1-2 & 18.6\% & 37.4\% & 14.7\% & 40.8\% & 18.5\% & 28.9\% \\
            2-3 & 18.1\% & 35.4\% & 20.8\% & 34.2\% & 17.3\% & 32.1\% \\
            3-4 & 21.3\% & 32.7\% & 14.8\% & 35.9\% & 16.8\% & 31.5\% \\
            4-5 & 23.7\% & 42.1\% & 20.1\% & 35.1\% & 21.2\% & 27.5\% \\
            5-6 & 23.4\% & 34.8\% & 18.8\% & 28.8\% & 19.0\% & 25.9\% \\
            \bottomrule
        \end{tabular}
    }
\end{table}

\begin{figure}[t]
    \centering
    \includegraphics[width=0.5\textwidth]{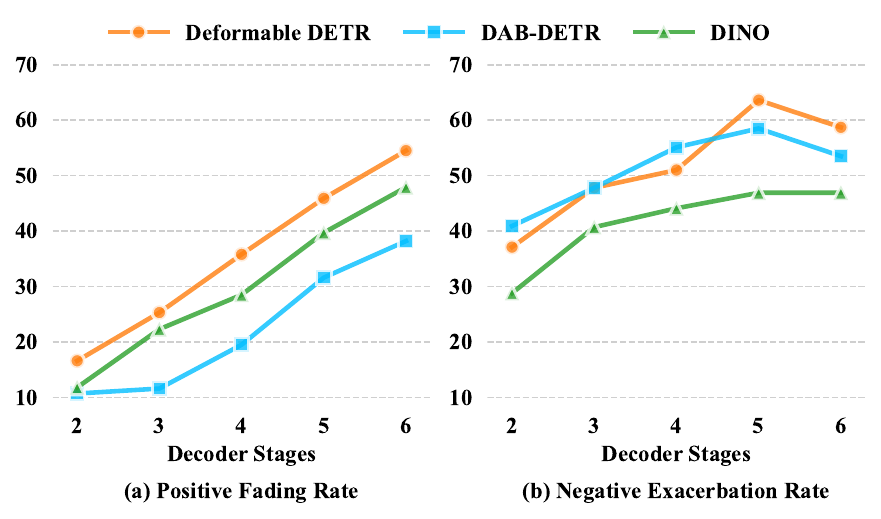}
    \caption{(a) Positive fading rates and (b) negative exacerbation rates across different decoder stages for various DETR models. All results are measured during the training phase.}
    \label{fig:fig4}
\end{figure}

\subsection{Motivation}
Previous work~\cite{chen2023enhanced} has observed that the final layer may not yield the best predictions from the query perspective and has utilized intermediate queries to improve detector training. However, its analysis focuses on query recollection for detector optimization rather than on how stage-wise prediction variation affects the quality of teacher supervision in DETR distillation. In this subsection, we analyze prediction quality across decoder stages and examine its implications for distillation targets.

Owing to DETRs' cascading update mechanism, both positive and negative updates are propagated sequentially across layers. Consequently, once a worse prediction occurs, its adverse effects may accumulate and become amplified in subsequent layers. To quantify this phenomenon, we calculate the error propagation rate at each layer to assess the extent of negative propagation. Specifically, for positive predictions, if the corresponding prediction in the next stage has a lower IoU and confidence score, it is identified as an error-propagated positive; for negative predictions, if the next stage prediction predicts the same category but with a higher confidence score, it is identified as an error-propagated negative. We compute these rates for Deformable DETR, DAB-DETR, and DINO. As presented in Table \ref{tab:tab1}, error propagation is widespread across all layers, with negative samples exhibiting particularly pronounced effects.

To further validate this trend, we extend the positive fading rate and negative exacerbation rate metrics proposed in \cite{chen2023enhanced} to all layers of the above models. As shown in Fig. \ref{fig:fig4}, each stage exhibits varying degrees of positive predictions degradation and negative predictions amplification, with both rates progressively increasing across stages. In the final layer, Deformable DETR achieves a positive fading rate of 50\%, DAB-DETR achieves 38\%, and DINO achieves 47.9\%. For the negative exacerbation rate, all models approach 50\%. These results confirm that non-monotonic prediction behavior is prevalent across all layers of DETR models and becomes increasingly pronounced as the stage increases.

Since knowledge distillation critically depends on the quality of teacher supervision, these observations indicate that relying solely on the current-stage predictions may provide suboptimal and even misleading guidance. In contrast, earlier decoder stages often contain more reliable predictions that are currently ignored. This motivates us to exploit superior intermediate predictions to refine teacher supervision before distillation, which forms the basis of our proposed Teacher Prediction Refinement Distillation.

\subsection{Teacher Prediction Refinement Distillation}
In this subsection, we detail the procedure of Teacher Prediction Refinement Distillation. An overview of our approach is illustrated in Fig.~\ref{fig:fig3}, and the PyTorch-like style pseudocode is shown in Algorithm~\ref {algo:stage-opt}. 
\subsubsection{Identifying Better Predictions}
As observed in the previous analysis, predictions from the current stage are sometimes degraded. Directly using them as supervision may therefore introduce noise or instability into the distillation process. Therefore, effectively exploiting better predictions from earlier stages becomes crucial for improving supervision quality. Accordingly, the first step is to identify better predictions from previous stages.

\definecolor{commentcolor}{RGB}{110,154,155}
\newcommand{\PyComment}[1]{\ttfamily\textcolor{commentcolor}{\# #1}}
\newcommand{\PyCode}[1]{\ttfamily\textcolor{black}{#1}}

\begin{algorithm}[t]
\PyComment{all\_outputs: outputs of all decoder stages} \\
\PyComment{positives: positive indices set} \\
\PyComment{negatives: negative indices set} \\
\PyComment{t: current stage index} \\

\PyCode{logits = all\_outputs["pred\_logits"][:t]} \\
\PyCode{boxes = all\_outputs["pred\_boxes"][:t]} \\
\PyCode{opt\_logits = logits[t-1].clone()} \\
\PyCode{opt\_boxes = boxes[t-1].clone()} \\

\PyComment{Positive Prediction Correction} \\
\PyCode{for (q, gt) in positives:} \\
\hspace*{1em}\PyCode{cls\_s = sigmoid(logits[:, q, gt\_cls])} \\
\hspace*{1em}\PyCode{iou\_s = IoU(boxes[:, q], gt\_box)} \\
\hspace*{1em}\PyCode{better = (iou\_s > iou\_s[t-1]) \& (cls\_s > cls\_s[t-1])} \\
\hspace*{1em}\PyCode{if better.any():} \\
\hspace*{2.5em}\PyCode{best = argmax(iou\_s[better] ** 0.75 * cls\_s[better] ** 0.25)} \\
\hspace*{2.5em}\PyCode{opt\_boxes[q] = boxes[best, q]} \\
\hspace*{2.5em}\PyCode{opt\_logits[q, gt\_cls] = logits[best, q, gt\_cls]} \\

\PyComment{Negative Prediction Suppression} \\
\PyCode{scores, cls = torch.max(logits, dim=-1)} \\
\PyCode{for q in negatives:} \\
\hspace*{1em}\PyCode{same\_cls = (cls[:, q] == cls[t-1, q])} \\
\hspace*{1em}\PyCode{lower = (scores[:, q] < scores[t-1, q])} \\
\hspace*{1em}\PyCode{better = same\_cls \& lower} \\
\hspace*{1em}\PyCode{if better.any():} \\
\hspace*{2.5em}\PyCode{best = argmin(scores[better, q])} \\
\hspace*{2.5em}\PyCode{opt\_logits[q, cls[t-1, q]] = logits[best, q, cls[t-1, q]]} \\

\PyCode{return \{"logits": opt\_logits, "boxes": opt\_boxes\}}
\caption{Pseudo code of TPRD in PyTorch-like style}
\label{algo:stage-opt}
\end{algorithm}

\noindent \textbf{Positive Prediction Correction.} Positive predictions at the current stage may not have the best IoU and classification scores. To enhance supervision quality, we aim to refine such degraded predictions by leveraging more reliable counterparts from earlier stages.

Therefore, A prediction from a previous stage is considered better when it has both a higher IoU and a classification score:
\begin{equation}
\begin{aligned}
\label{eq:6}
\mathcal{C}^{+}(\mathbf{p}_i^j,\mathbf{p}_i) = & \ \operatorname{IoU}(\mathbf{b}_i^j,\mathbf{b}^{gt}_{\hat{\sigma}(i)}) > \operatorname{IoU}(\mathbf{b}_i,\mathbf{b}^{gt}_{\hat{\sigma}(i)}) \\
 & \text{and } (\mathbf{c}_i^j)_{true} > (\mathbf{c}_i)_{true}.
\end{aligned}
\end{equation}
Here, $i \in \mathbf{I}^+$, $j$ represents the previous stage index, and ``true" represents the true class index. We then compare all previous stages with the current stage $k$ to identify better predictions as candidates. We store the corresponding stage indices instead:
\begin{equation}
    \label{eq:7}
    \mathbf{J}^{+}_{i} = \{ j \ | \ \mathcal{C}^+(\mathbf{p}_i^j,\mathbf{p}_i)=1 , \ j=1,\cdots,k-1 \}.
\end{equation}
Ideally, we would select the candidate with both the highest IoU and classification scores as the best candidate, but such cases are rare in practice. Therefore, we adopt a more practical strategy by choosing the candidate with the highest quality score\cite{chen2021disentangle} as the best candidate:
\begin{equation}
    \label{eq:8}
    \hat{j}_i^+ = \mathop{\arg\max}_{j \in \mathbf{J}^{+}_i} \ (\mathbf{c}_i^j)^{\gamma}_{true} \times \operatorname{IoU}(\mathbf{b}_i^j,\mathbf{b}^{gt}_{\hat{\sigma}(i)})^{1-\gamma},
\end{equation}
where $\gamma$ is a hyperparameter that balances the contributions of the IoU and classification scores, we set $\gamma = 0.25$ in our experiments. 

\noindent \textbf{Negative Prediction Suppression.} Negative predictions at the current stage may sometimes yield higher classification confidences than those from previous stages, introducing noisy supervision signals. To alleviate this issue, we refine such overconfident negative predictions by suppressing their excessive confidence. Therefore, a negative prediction from previous stages is regarded as better if it predicts the same class but with a lower confidence:
\begin{equation}
\begin{aligned}
\label{eq:9}
\mathcal{C}^{-}(\mathbf{p}_i^j,\mathbf{p}_i) = & \ \arg\max(\mathbf{c}_i^j) == \arg\max(\mathbf{c}_i) \\
 & \& \max(\mathbf{c}_i^j) < \max(\mathbf{c}_i),
\end{aligned}
\end{equation}
where $i \in \mathbf{I}^-$. We can then similarly obtain a candidate set $\mathbf{J}_i^{-}$ and select the prediction with the lowest classification score as the best candidate:
\begin{equation}
    \label{eq:10}
    \hat{j}_i^- = \mathop{\arg\min}_{j \in \mathbf{J}^{-}_i} \ max(\mathbf{c}_i^j).
\end{equation}
For simplicity, we omit the subscript $i$ and the superscripts $+$ and $-$ in the remainder of this section. The positive/negative partition is determined independently at each stage by its own Hungarian matching result, while cross-stage predictions are aligned only by query index. No cross-stage matching or re-partitioning is performed. If no earlier-stage prediction satisfies Eq. (\ref{eq:6}) or Eq. (\ref{eq:9}), the candidate set is empty, and the current-stage prediction is kept unchanged without replacement.

\subsubsection{Refining Teacher Predictions}
After identifying the best candidates from previous stages, the next key step is to effectively utilize them to refine the teacher’s predictions used for distillation. A straightforward approach is to directly adopt the best candidates from the previous stages as the refined ones. While this is feasible for bounding boxes, it completely overwrites the dark knowledge encoded in the current stage’s logits, thereby impairing effective knowledge transfer from the teacher model, which is empirically validated in Table~\ref{tab:tab6}.

\noindent \textbf{Maximum Dark Knowledge Preservation}. Therefore, as shown in Fig.~\ref{fig:fig3}(b), to preserve as much dark knowledge as possible, we perform a selective refinement by only replacing the logit of the target class while leaving all other logits unchanged. Formally, the replacement operation is defined as:
\begin{equation}
\label{eq:11}
\mathcal{R}(\mathbf{p}_i,\mathbf{p}_i^{\hat{j}}) = 
\begin{cases}
\mathbf{b}_i = \mathbf{b}_i^{\hat{j}},(\mathbf{z}_i)_t = (\mathbf{z}_i^{\hat{j}})_t, & i \in \mathbf{I}^+ \\
(\mathbf{z}_i)_t = (\mathbf{z}_i^{\hat{j}})_t, & i \in \mathbf{I}^-,
\end{cases}
\end{equation}
where $t$ denotes the target class index, $\mathbf{z}$ is the logits before activation. The refined teacher prediction for the current stage at index $i$ is then constructed as:
\begin{equation}
\label{eq:12}
\hat{\mathbf{p}}_i = \mathcal{R}(\mathbf{p}_i, \mathbf{p}_i^{\hat{j}}),
\end{equation}
yielding a set of refined teacher predictions $\{\hat{\mathbf{p}}_i = (\hat{\mathbf{c}}_i,\hat{\mathbf{b}}_i)\}_{i=1}^N$, which are used to replace the original predictions for subsequent distillation.

Although only the target-class logit is replaced, this modification may still slightly perturb inter-class relationships, thereby affecting dark knowledge to some extent. To further mitigate this issue, we adopt the Decoupled Knowledge Distillation (DKD)~\cite{zhao2022decoupled}. DKD decouples the distillation of target and non-target classes, maintaining inter-class dark knowledge among non-target logits while leveraging the refined target-class probability for supervision. The loss is defined as:
\begin{equation}
\label{eq:13}
\begin{aligned}
    \mathcal{L}_{DKL} = & \alpha \space [\hat{\mathbf{c}}_tlog\frac{\hat{\mathbf{c}}_t}{\mathbf{c}_t^S} + (1-\hat{\mathbf{c}}_t)log\frac{1-\hat{\mathbf{c}}_t}{1-\mathbf{c}_t^S}] \\
    & + \beta \sum_{i \neq t}^C \tilde{\mathbf{c}}_i^T log\frac{\tilde{\mathbf{c}}_i^T}{\tilde{\mathbf{c}}_i^S},
\end{aligned}
\end{equation}
where $\alpha$ and $\beta$ are hyperparameters balancing the two terms, and we set $\alpha = 1,\beta = 1$ in our experiments. The subscript $t$ denotes the target class, and superscripts $T$ and $S$ denote the teacher and student, respectively. $\tilde{\mathbf{c}}$ represents the softmax-normalized probabilities excluding the target class. Unlike the original definition in~\cite{zhao2022decoupled}, which defines the target class only for positive samples, our formulation extends it by treating the predicted class of negative samples as their target class. This generalization enables consistent optimization across both positive and negative predictions, thereby preserving comprehensive dark knowledge.

\subsection{Total Loss}
Finally, we define the total loss of our knowledge distillation framework as:
\begin{equation}
\label{eq:14}
\begin{aligned}
    \mathcal{L}_{total} = &\space  \mathcal{L}_{train} + \mathcal{L}_{specific} \\
    & + \sum_i^N [\lambda_{cls}\mathcal{L}_{DKL}(\hat{\mathbf{c}}_i,\mathbf{c}_i^S) + \lambda_{box}\mathcal{L}_{box} (\hat{\mathbf{b}}_i,\mathbf{b}_i^S)],
\end{aligned}
\end{equation}
where $\mathcal{L}_{train}$ is the standard training loss of DETR models. Notice that our method is a plug-and-play component that plugs in other methods, so the $\mathcal{L}_{specific}$ represents the additional loss used in their specific distillation methods.

\section{Experiments}
\subsection{Experimental Setup}
\subsubsection{Datasets} All experiments are conducted on two widely-used object detection benchmarks: MS COCO 2017~\cite{lin2014microsoft} and Pascal VOC~\cite{everingham2010pascal,everingham2015pascal}. The MS COCO 2017 dataset contains approximately 117K training images and 5K validation images spanning 80 object categories. Following the COCO protocol, we use mean Average Precision (mAP) as the primary evaluation metric.

For the Pascal VOC dataset, we follow the standard VOC 2007+2012 training setup and evaluate the models on the VOC 2007 test set, which consists of 20 object categories. We report $AP_{50}$ as the evaluation metric, which is the commonly adopted protocol in prior works.

\subsubsection{Implementation details} Our implementation is built upon Detrex~\cite{ren2023detrex}. Since our approach is designed as a plug-and-play module, all hyperparameters, optimization settings, and training schedules strictly follow those of the respective baseline frameworks to ensure fair comparison. All results reported are based on our reproduced experiments for fair comparison. All experiments are conducted on NVIDIA A800 GPUs with a total batch size of 16.

\begin{table*}[ht]
\centering
\caption{Results of our proposed TPRD on the MS COCO 2017 dataset. We compare with different SOTA DETR distillation methods on different DETR variants. Values are reported as mean $\pm$ std over multiple random seeds.}

\newcommand{\highlightcell}[1]{\cellcolor{gray!25}\textbf{#1}}

\begin{tabular}{cccccccccc}
\toprule
Teacher/Student & Methods & AP & $AP_{50}$ & $AP_{75}$ &
$AP_s$ & $AP_M$ & $AP_L$ & GFLOPs & FPS \\
\midrule

\multicolumn{10}{c}{\textbf{Deformable DETR\cite{zhu2020deformable}}} \\
\midrule

\multirow{12}{*}{\begin{tabular}{c}
ResNet-50 (40.1M) \\
to \\
ResNet-18 (23.6M)
\end{tabular}}
& Teacher & 44.7&63.6&48.7&27.1&47.6&59.6 & 171 & 20 \\
& Student &40.1&58.1&43.7&22.4&42.8&54.2 & 127 & 25 \\
\cmidrule{2-10}
& QSKD\cite{liu2024knowledge} &40.7$_{\scriptsize\pm0.13}$&58.2$_{\scriptsize\pm0.27}$&44.3$_{\scriptsize\pm0.17}$&23.0$_{\scriptsize\pm0.32}$&43.7$_{\scriptsize\pm0.35}$&54.5$_{\scriptsize\pm0.18}$ & 127 & 25 \\
& + Ours & \highlightcell{41.4$_{\scriptsize\pm0.18}$} & \highlightcell{59.4$_{\scriptsize\pm0.29}$}
& \highlightcell{45.0$_{\scriptsize\pm0.23}$} & \highlightcell{23.1$_{\scriptsize\pm0.40}$}
& \highlightcell{43.9$_{\scriptsize\pm0.28}$} & \highlightcell{56.0$_{\scriptsize\pm0.32}$} & 127 & 25 \\
\cmidrule{2-10}
& DETRDistill\cite{chang2023detrdistill} &41.9$_{\scriptsize\pm0.13}$&59.2$_{\scriptsize\pm0.23}$&45.6$_{\scriptsize\pm0.20}$&24.3$_{\scriptsize\pm0.43}$&45.0$_{\scriptsize\pm0.38}$&55.6$_{\scriptsize\pm0.33}$ & 127 & 25 \\
& + Ours & \highlightcell{42.6$_{\scriptsize\pm0.22}$} & \highlightcell{60.4$_{\scriptsize\pm0.19}$}
& \highlightcell{46.4$_{\scriptsize\pm0.22}$} & \highlightcell{24.7$_{\scriptsize\pm0.54}$}
& \highlightcell{45.2$_{\scriptsize\pm0.21}$} & \highlightcell{56.4$_{\scriptsize\pm0.31}$} & 127 & 25 \\
\cmidrule{2-10}
& $\text{D}^3$ETR\cite{ChenD3ETR} &42.1$_{\scriptsize\pm0.26}$&60.1$_{\scriptsize\pm0.24}$&45.8$_{\scriptsize\pm0.16}$&23.6$_{\scriptsize\pm0.36}$&45.2$_{\scriptsize\pm0.28}$&56.7$_{\scriptsize\pm0.33}$ & 127 & 25\\
& + Ours & \highlightcell{42.7$_{\scriptsize\pm0.18}$} & \highlightcell{60.6$_{\scriptsize\pm0.19}$}
& \highlightcell{46.4$_{\scriptsize\pm0.25}$} & \highlightcell{23.7$_{\scriptsize\pm0.29}$}
& \highlightcell{45.8$_{\scriptsize\pm0.19}$} & \highlightcell{57.0$_{\scriptsize\pm0.16}$} & 127 & 25 \\
\cmidrule{2-10}
& KD-DETR\cite{wang2024kd} &43.3$_{\scriptsize\pm0.18}$&61.2$_{\scriptsize\pm0.26}$&47.3$_{\scriptsize\pm0.30}$&25.4$_{\scriptsize\pm0.25}$&46.7$_{\scriptsize\pm0.38}$&56.9$_{\scriptsize\pm0.32}$ & 127 & 25 \\
& + Ours & \highlightcell{43.7$_{\scriptsize\pm0.20}$} & \highlightcell{61.6$_{\scriptsize\pm0.21}$}
& \highlightcell{47.6$_{\scriptsize\pm0.26}$} & \highlightcell{25.6$_{\scriptsize\pm0.43}$}
& \highlightcell{46.9$_{\scriptsize\pm0.35}$} & \highlightcell{57.8$_{\scriptsize\pm0.30}$} & 127 & 25 \\
\cmidrule{2-10}
& CloCKDistill\cite{lan2025clockdistill} &43.2$_{\scriptsize\pm0.14}$&61.2$_{\scriptsize\pm0.21}$&47.3$_{\scriptsize\pm0.27}$&25.0$_{\scriptsize\pm0.37}$&46.2$_{\scriptsize\pm0.37}$&57.4$_{\scriptsize\pm0.21}$ & 127 & 25 \\
& + Ours & \highlightcell{44.0$_{\scriptsize\pm0.19}$} & \highlightcell{62.1$_{\scriptsize\pm0.26}$}
& \highlightcell{48.0$_{\scriptsize\pm0.33}$} & \highlightcell{26.2$_{\scriptsize\pm0.56}$}
& \highlightcell{47.1$_{\scriptsize\pm0.25}$} & \highlightcell{57.7$_{\scriptsize\pm0.29}$} & 127 & 25 \\

\midrule

\multirow{4}{*}{\begin{tabular}{c}
ResNet-50 (40.1M)\\
to \\
MobileNetV2 (16.5M)
\end{tabular}}
& Teacher &44.7&63.6&48.7&27.1&47.6&59.6 & 171 & 20 \\
& Student &41.1&60.0&44.6&23.1&44.1&55.5 & 101 & 22 \\
\cmidrule{2-10}
& KD-DETR\cite{wang2024kd} &43.3$_{\scriptsize\pm0.19}$&62.0$_{\scriptsize\pm0.16}$&47.2$_{\scriptsize\pm0.23}$&25.1$_{\scriptsize\pm0.53}$&46.6$_{\scriptsize\pm0.37}$&57.6$_{\scriptsize\pm0.33}$ & 101 & 22 \\
& + Ours & \highlightcell{44.0$_{\scriptsize\pm0.22}$} & \highlightcell{62.3$_{\scriptsize\pm0.25}$}
& \highlightcell{48.1$_{\scriptsize\pm0.23}$} & \highlightcell{25.9$_{\scriptsize\pm0.27}$}
& \highlightcell{47.1$_{\scriptsize\pm0.21}$} & \highlightcell{58.6$_{\scriptsize\pm0.32}$} & 101 & 22 \\

\midrule

\multirow{4}{*}{\begin{tabular}{c}
ResNet-101 (59.0M)\\
to \\
ResNet-50 (40.1M)
\end{tabular}}
& Teacher &47.6&67.3&51.7&29.4&51.6&62.8 & 238 & 17 \\
& Student &44.7&63.5&48.7&27.1&47.6&59.6 & 171 & 20 \\
\cmidrule{2-10}
& KD-DETR\cite{wang2024kd} &46.2$_{\scriptsize\pm0.16}$&65.3$_{\scriptsize\pm0.17}$&50.6$_{\scriptsize\pm0.15}$&27.6$_{\scriptsize\pm0.35}$&49.4$_{\scriptsize\pm0.31}$&61.4$_{\scriptsize\pm0.17}$ & 171 & 20 \\
& + Ours & \highlightcell{47.1$_{\scriptsize\pm0.20}$} & \highlightcell{66.2$_{\scriptsize\pm0.23}$}
& \highlightcell{51.8$_{\scriptsize\pm0.16}$} & \highlightcell{28.6$_{\scriptsize\pm0.54}$}
& \highlightcell{50.0$_{\scriptsize\pm0.23}$} & \highlightcell{62.3$_{\scriptsize\pm0.27}$} & 171 & 20 \\

\midrule

\multicolumn{10}{c}{\textbf{DAB-DETR\cite{liu2022dab}}} \\
\midrule

\multirow{4}{*}{\begin{tabular}{c}
ResNet-50 (43.7M)\\
to \\
ResNet-18 (31.0M)
\end{tabular}}
& Teacher &43.3&63.1&44.7&21.5&45.7&62.3 & 89 & 30 \\
& Student & 36.2&56.1&37.9&16.9&39.0&53.5 & 48 & 36 \\
\cmidrule{2-10}
& KD-DETR\cite{wang2024kd} & 41.4$_{\scriptsize\pm0.20}$&61.4$_{\scriptsize\pm0.27}$&43.9$_{\scriptsize\pm0.20}$&20.4$_{\scriptsize\pm0.51}$&44.7$_{\scriptsize\pm0.37}$&61.0$_{\scriptsize\pm0.24}$ & 48 & 36 \\
& + Ours & \highlightcell{41.8$_{\scriptsize\pm0.27}$} & \highlightcell{61.5$_{\scriptsize\pm0.22}$}
& \highlightcell{44.5$_{\scriptsize\pm0.25}$} & \highlightcell{20.7$_{\scriptsize\pm0.52}$}
& \highlightcell{45.3$_{\scriptsize\pm0.35}$} & \highlightcell{61.5$_{\scriptsize\pm0.17}$} & 48 & 36\\

\midrule

\multicolumn{10}{c}{\textbf{DINO\cite{zhang2022dino}}} \\
\midrule

\multirow{4}{*}{\begin{tabular}{c}
ResNet-50 (47.7M)\\
to \\
ResNet-18 (31.1M)
\end{tabular}}
& Teacher & 49.9&69.0&55.3&34.6&54.1&64.6 & 245 & 15 \\
& Student &44.0&61.2&48.1&27.4&46.9&56.9 & 200 & 18 \\
\cmidrule{2-10}
& KD-DETR\cite{wang2024kd} &48.0$_{\scriptsize\pm0.26}$&65.1$_{\scriptsize\pm0.24}$&52.3$_{\scriptsize\pm0.24}$&30.5$_{\scriptsize\pm0.45}$&50.7$_{\scriptsize\pm0.28}$&62.0$_{\scriptsize\pm0.26}$ & 200 & 18 \\
& + Ours & \highlightcell{48.5$_{\scriptsize\pm0.16}$} & \highlightcell{65.3$_{\scriptsize\pm0.18}$}
& \highlightcell{52.9$_{\scriptsize\pm0.20}$} & \highlightcell{31.3$_{\scriptsize\pm0.40}$}
& \highlightcell{50.9$_{\scriptsize\pm0.28}$} & \highlightcell{62.4$_{\scriptsize\pm0.21}$} & 200 & 18 \\

\midrule

\multicolumn{10}{c}{\textbf{RT-DETR\cite{zhao2024detrs}}} \\
\midrule

\multirow{4}{*}{\begin{tabular}{c}
ResNet-50 (36M)\\
to \\
ResNet-18 (20M)
\end{tabular}}
& Teacher &51.3&69.6&55.5&33.7&56.0&69.2 & 100 & 58 \\
& Student &45.9&63.1&49.8&28.3&49.7&62.1 & 60 & 90 \\
\cmidrule{2-10}
& KD-DETR\cite{wang2024kd} &46.2$_{\scriptsize\pm0.27}$&63.3$_{\scriptsize\pm0.18}$&49.9$_{\scriptsize\pm0.22}$&28.1$_{\scriptsize\pm0.50}$&49.6$_{\scriptsize\pm0.26}$&63.8$_{\scriptsize\pm0.25}$ & 60 & 90 \\
& + Ours & \highlightcell{46.9$_{\scriptsize\pm0.14}$} & \highlightcell{63.9$_{\scriptsize\pm0.30}$}
& \highlightcell{50.6$_{\scriptsize\pm0.33}$} & \highlightcell{28.6$_{\scriptsize\pm0.39}$}
& \highlightcell{50.1$_{\scriptsize\pm0.28}$} & \highlightcell{64.5$_{\scriptsize\pm0.35}$} & 60 & 90\\

\bottomrule
\end{tabular}
\label{tab:tab3}
\end{table*}

\subsection{Comparison with DETR-based distillation methods}
\subsubsection{Evaluation on MS COCO} We evaluate our method across five representative DETR distillation frameworks: DETRDistill, KD-DETR, QSKD, $\text{D}^3$ETR, and CLoCKDistill, on four DETR variants: Deformable DETR, DAB-DETR, DINO, and RT-DETR. All five distillation frameworks are validated on Deformable DETR, while KD-DETR is additionally evaluated on DAB-DETR, DINO, and RT-DETR. To further evaluate the method under different teacher-student capacities, we also conduct experiments using a ResNet-50 teacher with a MobileNetV2 student and a ResNet-101 teacher with a ResNet-50 student. All results are averaged over three independent runs with different random seeds, and the standard deviation is reported.

As presented in Table \ref{tab:tab3}, our proposed method achieves consistent performance improvements across all DETR variants and distillation frameworks. In terms of distillation methods, TPRD steadily enhances the representative baselines. Specifically, on Deformable DETR with a ResNet-50 teacher and a ResNet-18 student, our method brings additional gains of +0.7 AP over both QSKD and DETRDistill, +0.6 AP over $\text{D}^3$ETR, +0.4 AP over KD-DETR, and +0.8 AP over CLoCKDistill. In terms of DETR architectures, our method also shows consistent gains across diverse variants, ranging from the single-scale DAB-DETR to the advanced DINO,  which improves KD-DETR on DAB-DETR, DINO, and RT-DETR by +0.4 AP, +0.5 AP, and +0.7 AP, respectively. In terms of different teacher-student capacities, TPRD improves KD-DETR from 43.3 AP to 44.0 AP when using a ResNet-50 teacher and a MobileNetV2 student, and from 46.2 AP to 47.1 AP when using a ResNet-101 teacher and a ResNet-50 student. These consistent improvements verify the effectiveness and robustness of our approach across the evaluated DETR variants and teacher–student configurations. By refining the teacher’s supervision signals, TPRD enables the student to learn from more accurate and stable guidance, thereby enhancing overall distillation efficiency across both frameworks and model architectures.

\begin{table*}[t]
    \centering
    \caption{Comparison with different CNN-based distillation methods on Deformable DETR. For the first three methods, feature-level distillation is applied using encoder outputs. For CrossKD and Ours, we implement them based on DETRDistill.}
    \label{tab:tab4}
        \begin{tabular}{ccccccccc}
            \toprule
            Models & Methods & Epochs & AP & $AP_{50}$ & $AP_{75}$ & $AP_S$ & $AP_M$ & $AP_L$ \\
            \toprule
            \multirow[=]{7}{*}{Deformable DETR\cite{zhu2020deformable}} & ResNet-50(T) & 50 & 44.7 & 63.6 & 48.7 & 27.1 & 47.6 & 59.6 \\
            & ResNet-18(S) & 50 & 40.1 & 58.1 & 43.7 & 22.4 & 42.8 & 54.2 \\
            \cmidrule{2-9}
            & FitNet\cite{romero2015fitnetshintsdeepnets} & 50 & 40.5 & 58.6 & 43.4 & 22.8 & 43.0 & 54.0 \\
            & FGD\cite{yang2022focal} & 50 & 40.8 & 58.7 & 44.0 & 23.1 & 43.5 & 54.2 \\
            & MGD\cite{yang2022masked} & 50 & 40.7 & 58.9 & 43.9 & 23.0 & 43.3 & 54.3 \\
            & CrossKD\cite{wang2023crosskd} & 50 & 41.4 & 58.7 & 45.2 & 23.5 & 44.2 & 54.4 \\
            & Ours & 50 & \textbf{42.6} & \textbf{60.4} & \textbf{46.4} & \textbf{24.7} & \textbf{45.2} & \textbf{56.4} \\
            \bottomrule
        \end{tabular}
\end{table*}

\begin{table}[t]
    \centering
    \caption{Results of our proposed TPRD on the Pascal VOC dataset. We compare with different SOTA DETR distillation methods on Deformable DETR.}
    \label{tab:voc}
        \begin{tabular}{cccc}
            \toprule
            Models & Methods & Epochs & $AP_{50}$ \\
            \toprule
            \multirow[=]{10}{*}{Deformable DETR\cite{zhu2020deformable}} & ResNet-50(T) & 50 & 68.2 \\
            & ResNet-18(S) & 50 & 63.0 \\
            \cmidrule{2-4}
            & QSKD\cite{liu2024knowledge} & 50 & 72.5  \\
            & + Ours & 50 & \textbf{72.9(+0.4)} \\
            \cmidrule{2-4}
            & DETRDistill\cite{chang2023detrdistill} & 50 & 69.6 \\
            & + Ours & 50 & \textbf{70.9(+1.3)}  \\
            \cmidrule{2-4}
            & $\text{D}^3$ETR\cite{ChenD3ETR} & 50 & 69.7 \\
            & + Ours & 50 & \textbf{71.9(+2.2)} \\
            \cmidrule{2-4}
            & KD-DETR\cite{wang2024kd} & 50 & 71.0 \\
            & + Ours & 50 & \textbf{71.9(+0.9)} \\
            \bottomrule
        \end{tabular}
\end{table}

\subsubsection{Evaluation on Pascal VOC} To further validate the effectiveness of our proposed method on different datasets, we conduct additional experiments on the Pascal VOC benchmark. We follow the standard training protocol on VOC, and all the hyperparameter settings are the same as those used in MS COCO. 

As shown in Table~\ref{tab:voc}, our method consistently improves the performance of various DETR-based distillation frameworks on Pascal VOC. Specifically, when integrated with QSKD, our approach further improves the student performance from 72.5 to 72.9 $AP_{50}$, yielding a gain of +0.4. More notably, larger improvements are observed on DETRDistill and $\text{D}^3$ETR, where our method brings additional gains of +1.3 and +2.2 $AP_{50}$, respectively. Even when combined with the strong KD-DETR baseline, our method still achieves a consistent improvement of +0.9 $AP_{50}$. These results demonstrate that our method generalizes well across datasets. On Pascal VOC, where models typically converge faster and predictions are more sensitive to supervision quality, refining the teacher’s predictions remains highly effective.


\subsection{Comparison with CNN-based distillation methods}
We compare our method with several representative CNN-based distillation methods to further validate its advantages, including the feature-based methods FitNet~\cite{romero2015fitnetshintsdeepnets}, FGD~\cite{yang2022focal}, and MGD~\cite{yang2022masked}, as well as the logits-based method CrossKD~\cite{wang2023crosskd}. All methods are reproduced using their official implementations to ensure fairness. For the feature-based approaches, encoder outputs are used for feature-level distillation, while CrossKD is implemented based on DETRDistill by treating the encoder as the neck and the decoder as the head.

As reported in Table~\ref{tab:tab4}, CNN-based distillation methods exhibit limited improvements when directly applied to DETR architectures. This can be attributed to their reliance on dense feature alignment and their failure to directly distill decoder predictions, which are more closely related to the final detection results. In contrast, our method achieves substantial performance gains, outperforming the best feature-based method by +1.8 AP and surpassing CrossKD by +1.2 AP. These results demonstrate that traditional CNN-based distillation strategies struggle to transfer knowledge effectively in DETRs, further underscoring the need to design distillation techniques tailored for transformer-based detectors.

\begin{table}
    \caption{Effectiveness of each component in our method. DETRDistill serves as the baseline when all the components are removed.}
     \centering
     \label{tab:tab5}
        \begin{tabular}{ccccccc}
            \toprule
            PPC & NPS & MDKP & AP & $AP_s$ & $AP_m$ & $AP_L$  \\
            \toprule
            &  &  & 41.9 & 24.3 & 45.0 & 55.6 \\
            $\checkmark$ & & & 41.8 & 24.1 & 44.6 & 55.0 \\
            & $\checkmark$ & & 42.1 & 24.4 & 45.2 & 55.0 \\
            & & $\checkmark$ & 42.0 & 24.0 & 44.6 & 55.8 \\
            $\checkmark$ & $\checkmark$ & & 41.8 & 24.6 & 44.4 & 55.1 \\
            $\checkmark$ & & $\checkmark$ & 42.2 & 24.1 & 45.0 & 55.8 \\
            & $\checkmark$ & $\checkmark$ & 42.3 & 24.4 & \textbf{45.4} & 56.2 \\
            $\checkmark$ & $\checkmark$ & $\checkmark$ & \textbf{42.6} & \textbf{24.7} & 45.2 & \textbf{56.4} \\
             \bottomrule
        \end{tabular}
\end{table}

\subsection{Ablation Study}
In this section, we validate the effectiveness of each component and conduct some further analyses. All the experiments in this section are conducted on Deformable DETR with ResNet-50 as the teacher and ResNet-18 as the student. We train the student models under a 50-epoch training schedule.

\subsubsection{The effectiveness of each component} To evaluate the contribution of each component in our method, we conduct an ablation experiment on each module, as shown in Table \ref{tab:tab5}. Starting from the baseline using DETRDistill, the model achieves 41.9 AP. When MDKP is disabled, the improvements brought by PPC and NPS remain limited and can even become negative. Using PPC alone slightly reduces performance by 0.1 AP, while NPS yields only a modest gain of 0.2 AP. Even applying PPC and NPS together fails to deliver additional benefits. This suggests that, although these two modules refine the teacher’s predictions, the accompanying loss of dark knowledge weakens their effectiveness and may overshadow the advantages of refined supervision. However, once MDKP is introduced to counteract this issue, both combinations exhibit clear improvements: PPC+MDKP reaches 42.2 AP, and NPS+MDKP further increases performance to 42.3 AP. It is also worth noting that MDKP alone contributes only a minor 0.1 AP improvement, likely attributable to DKD\cite{zhao2022decoupled}, implying that the performance gains of TPRD cannot be explained simply by incorporating DKD. Finally, integrating all three components achieves the highest performance of 42.6 AP, reflecting their complementary roles: PPC and NPS enhance the quality of teacher predictions, while MDKP preserves the essential dark knowledge, allowing TPRD to construct more reliable and informative supervision signals.
\begin{table}[tbp]
    \centering
    \caption{Comparison results of different TPRD implementations.  “GT” denotes replacing current predictions with ground-truth values instead of superior stage predictions, while “RA” refers to replacing all logits at the current stage.}
    \label{tab:tab6}
        \begin{tabular}{c|cccc}
            \toprule
            Methods & AP & $AP_S$ & $AP_M$ & $AP_L$ \\
            \toprule
            Baseline & 41.9 & 24.3 & 45.0 & 55.6 \\
            GT & 42.2(+0.3) & 24.1 & 45.2 & 56.3 \\
            RA & 42.3(+0.4) & \textbf{24.8} & 45.1 & 55.8 \\
            Ours & \textbf{42.6(+0.7)} & 24.7 & \textbf{45.2} & \textbf{56.4} \\
            \bottomrule
        \end{tabular}
\end{table}

\begin{table}[t]
    \centering
    \caption{Distillation performance of applying TPRD at different starting stages.}
    \label{tab:tab7}
        \begin{tabular}{c|cccc}
            \toprule
            Start Stage & AP & $AP_S$ & $AP_M$ & $AP_L$ \\
            \toprule
            6 & 41.9 & 24.4 & 44.6 & 55.0 \\
            5 & 42.1 & 24.1 & 45.0 & 55.6 \\
            4 & 42.0 & 23.7 & 44.9 & 56.4 \\
            3 & 42.2 & 23.7 & 45.1 & 56.3 \\
            2 & \textbf{42.6} & \textbf{24.7} & \textbf{45.2} & \textbf{56.4} \\
            \bottomrule
        \end{tabular}
\end{table}

\subsubsection{Comparison of different TPRD implementations} In the MDKP module, we propose to replace only the target-class logit during the refinement process, rather than all logits. Moreover, we utilize superior predictions from earlier stages as replacements instead of directly using ground-truth values. To validate our choice, we conduct comparative experiments, and the results are reported in Table \ref{tab:tab6}. ``GT'' refers to replacing the current stage’s probabilities and bounding boxes with the ground-truth. ``RA'' refers to replacing all logits of the current stage. For GT replacement, we observe only a 0.3 AP improvement over the baseline. This may be because simply replacing with GTs overlaps with the supervision signals already provided by GTs, thereby reducing supervision diversity and learning dynamics. Replacing all logits yields a slightly larger gain of +0.4 AP over the baseline but is still inferior to our final method. This aligns with the analysis we mentioned above that full replacement erases the teacher model’s dark knowledge at the current stage, limiting its ability to transfer inter-class relations to the student model.

\subsubsection{Effect of different starting stages of TPRD} Previous analyses suggest that error propagation becomes increasingly amplified in deeper layers, which implies that earlier intervention may provide greater benefits. To examine this effect, we investigate the impact of applying TPRD from different starting stages. The results are shown in Table \ref{tab:tab7}. When TPRD is applied across all stages, the model achieves the highest performance. As the starting point moves deeper into the decoder, the gains gradually diminish, and enabling TPRD only at the final stage results in negligible improvement. This pattern indicates that early-stage refinement plays a crucial role: since TPRD refines the teacher’s predictions by selecting higher-quality outputs from all preceding layers, enabling it earlier allows more decoder layers to benefit from refined predictions.

\begin{table}[t]
    \centering
    \caption{Distillation performance with fewer encoders and decoders. When reducing the number of decoder stages, we follow the stage mapping strategy proposed in \cite{chang2023detrdistill}.}
    \label{tab:tab8}
        \begin{tabular}{cccccc}
            \toprule
            \#Enc./Dec. & Model & AP & $AP_S$ & $AP_M$ & $AP_L$ \\
            \toprule
            \multirow{2}{*}{6/3} & DETRDistill & 41.5 & \textbf{24.2} & 44.5 & 54.0 \\
            & Ours & \textbf{41.9(+0.4)} & 23.8 & \textbf{45.1} & \textbf{55.9} \\
            \midrule
            \multirow{2}{*}{3/6} & DETRDistill & 39.9 & 22.8 & 42.8 & 52.9 \\
            & Ours & \textbf{40.1(+0.2)} & \textbf{23.4} & \textbf{42.8} & \textbf{53.6} \\
            \midrule
            \multirow{2}{*}{3/3} & DETRDistill & 38.1 & \textbf{21.5} & 41.0 & 50.2 \\
            & Ours & \textbf{39.0(+0.9)} & 21.2 & \textbf{41.8} & \textbf{52.9} \\
            \bottomrule
        \end{tabular}
\end{table}

\begin{figure}[t]
    \centering
    \includegraphics[width=0.5\textwidth]{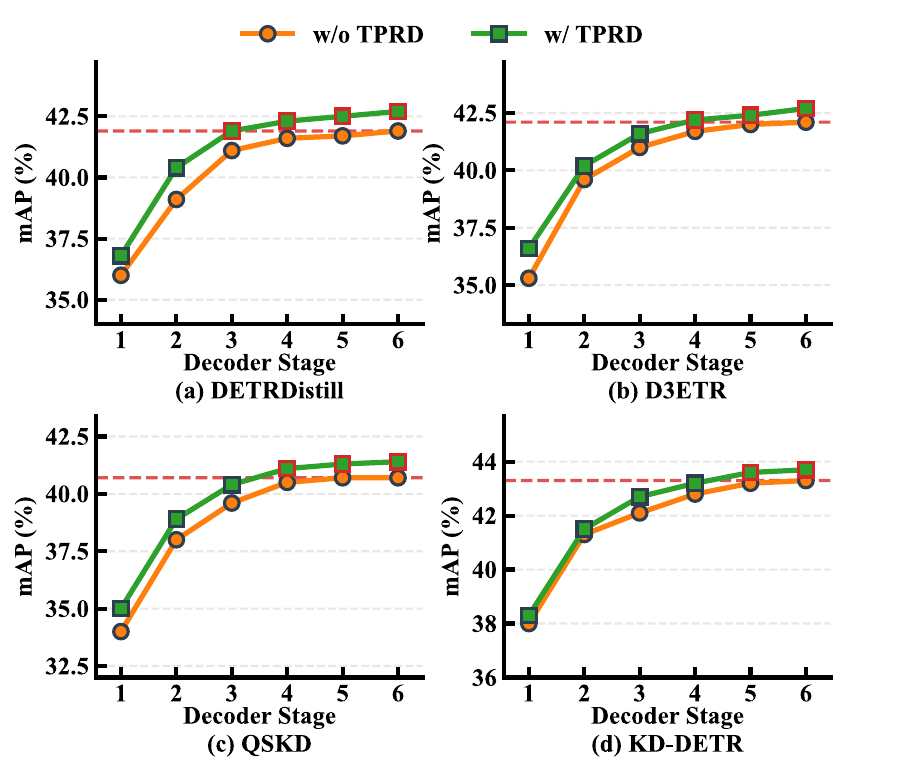}
    \caption{Inference performance of the student model across different stages on different DETR distillation methods. The red boxes highlight the stages that exceed the final-stage performance of the corresponding baseline methods.}
    \label{fig:fig8}
\end{figure}

\subsubsection{Distillation on fewer encoder and decoder stages} In practice, the number of encoder and decoder stages of the student model may not always match those in the teacher model. To investigate the robustness of our method under such conditions, we further conduct experiments by reducing the number of encoder and decoder stages of the student model. Reducing the decoder layers breaks the stage-to-stage correspondence between the teacher and student decoders, so we simply adopt the stage matching strategy proposed in \cite{chang2023detrdistill}. As shown in Table \ref{tab:tab8}, our method consistently improves performance even when the stage numbers differ. When only the encoder depth is reduced, the mAP increases from 39.9 to 40.1. When only the decoder depth is reduced, the gain becomes more pronounced, improving mAP from 41.5 to 41.9. Even under the most compressed setting, where both the encoder and decoder contain only three layers, TPRD still yields a notable improvement, increasing performance from 38.1 to 39.0.

\subsubsection{Distillation performance of student model across stages}
We analyze the stage-wise detection performance of the student model to understand how our method affects multi-stage knowledge transfer, as shown in Fig.~\ref{fig:fig8}. Across all distillation methods, TPRD delivers consistent improvements at every decoder stage, indicating that our method enhances supervision throughout the decoding process rather than only at the final layer. The performance gain is particularly evident in the middle-to-late stages, where prediction degradation typically accumulates. In DETRDistill, TPRD boosts the 3rd and 4th stages by +0.8 and +0.7 AP, respectively, while in QSKD and $\text{D}^3$ETR, the 3rd-stage improvements reach +0.8 and +0.6 AP. Additionally, we find an interesting fact that the student model trained with TPRD achieves comparable or even superior results using fewer decoder stages. For example, the 4th stage of KD-DETR (43.2 AP) and the 3rd stage of DETRDistill (42.3 AP) already match or surpass their respective full-depth baselines. Taken together, the improvements observed across stages indicate that TPRD not only improves the efficiency of knowledge transfer across stages but also enables potential model compression without sacrificing detection accuracy.

\begin{figure*}[t]
    \centering
    \includegraphics[width=0.8\textwidth]{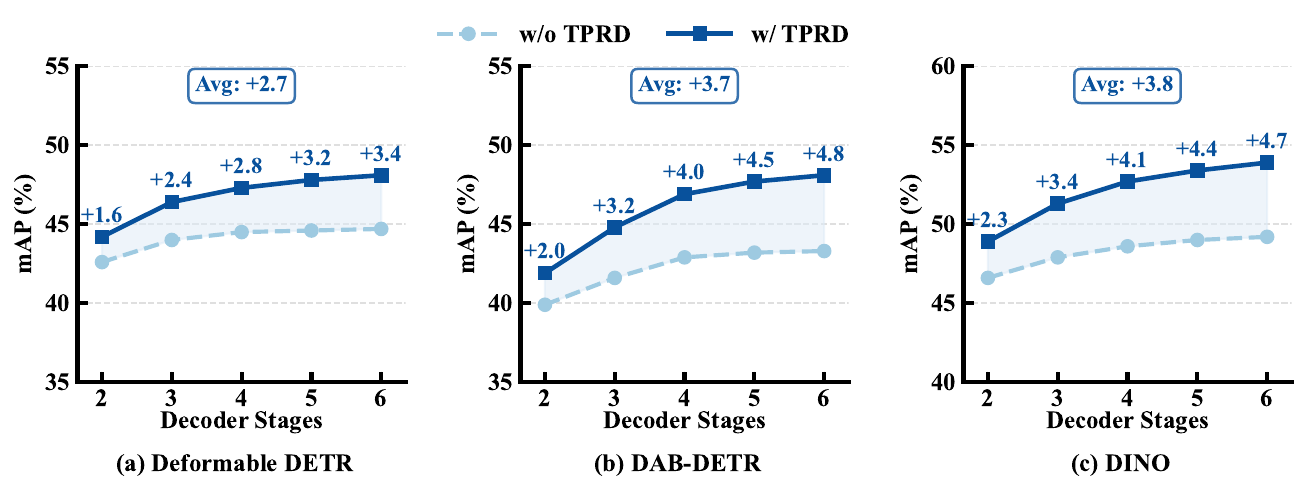}
    \caption{Quality of teacher supervision signals across different decoder stages before and after applying TPRD on different DETR variants. The dashed blue line shows the original predictions produced by the teacher model, while the solid blue line corresponds to the reconstructed supervision signals obtained by TPRD. Note that TPRD does not modify the teacher model parameters nor affect its inference behavior. The reported AP reflects the quality of supervision signals rather than the teacher’s inference performance.}
    \label{fig:fig7}
\end{figure*}

\begin{figure}[t]
    \centering
    \includegraphics[width=0.5\textwidth]{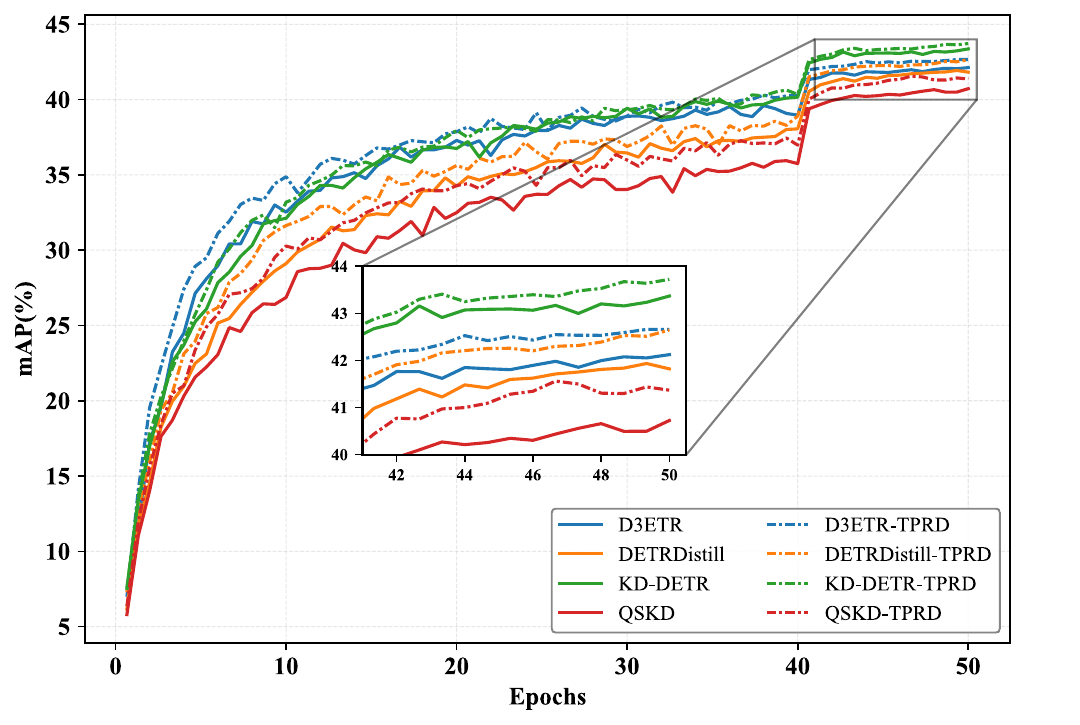}
    \caption{Training curves of different distillation methods with and without applying TPRD. The solid lines denote the original methods, while the dashed lines represent their TPRD versions. The inset highlights the final training phase, showing consistent performance gains and faster convergence brought by TPRD across all distillation frameworks.}
    \label{fig:fig10}
\end{figure}

\subsection{Additional analyses}
\subsubsection{Analysis of teacher supervision quality before and after applying TPRD}
TPRD serves as a supervision refinement mechanism that reconstructs higher-quality supervision signals from the teacher’s intermediate outputs. To evaluate the effectiveness of this refinement, we compare the supervision quality of each decoder stage before and after applying TPRD. As illustrated in Fig.~\ref{fig:fig7}, the reconstructed predictions consistently exhibit higher quality across all DETR variants and decoder stages. Since TPRD uses ground-truth annotations to select better predictions, and the same annotations are used to calculate AP, this analysis evaluates the quality of reconstructed supervision signals rather than the teacher model’s inference predictions.

For Deformable DETR, the average AP of the reconstructed supervision signals increases by +2.7 AP. Even larger gains are observed for DAB-DETR (+3.7 AP) and DINO (+3.8 AP). Notably, the improvements become more pronounced at deeper decoder stages. For example, the last stage of DINO improves from 49.2 AP to 53.9 AP after applying TPRD. This indicates that TPRD effectively mitigates the accumulation of prediction degradation along the decoding process, thereby providing more accurate and stable supervision signals for distillation.

\begin{figure*}[t]
    \centering
    \includegraphics[width=\textwidth]{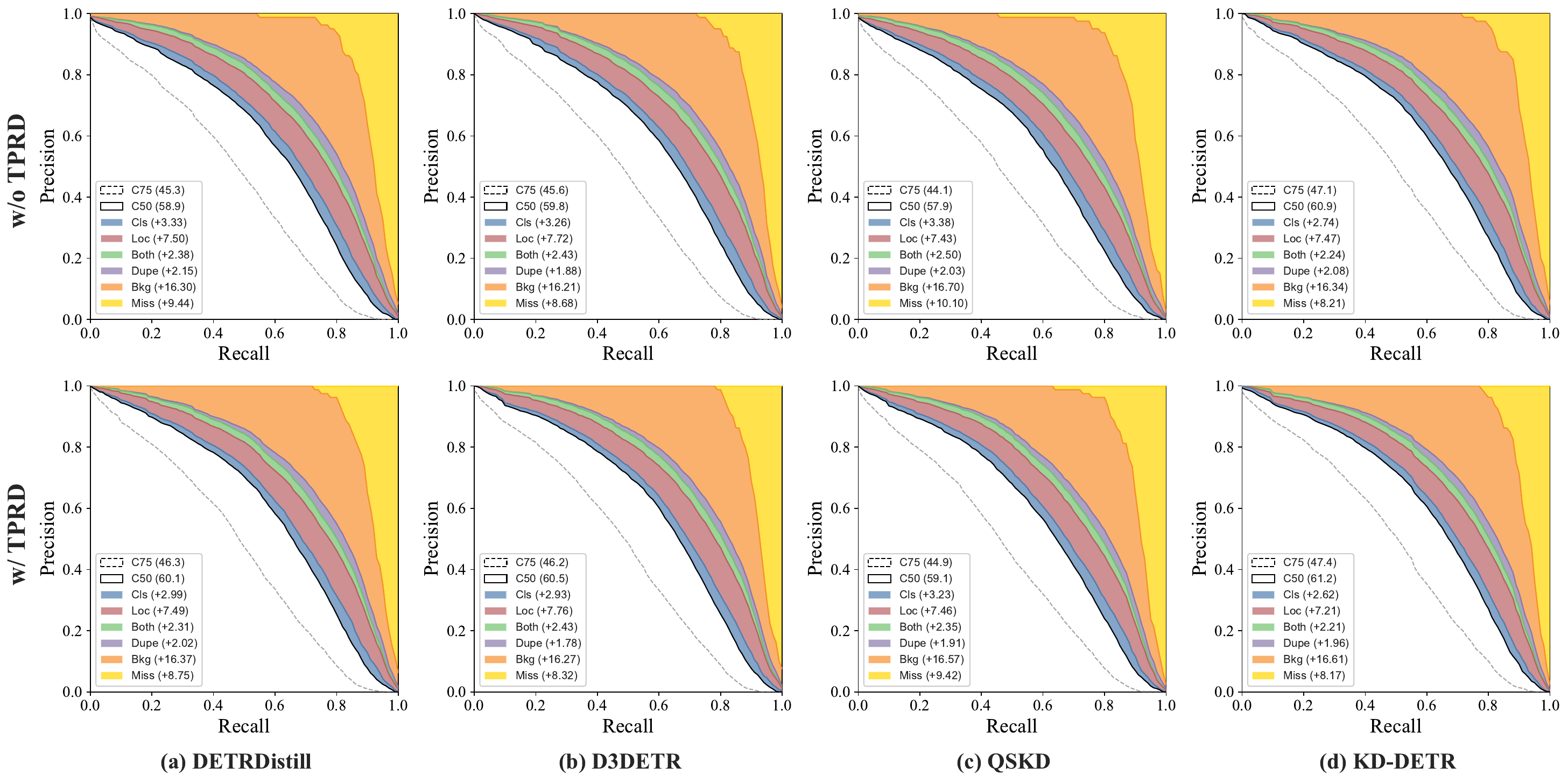}
    \caption{Error-type distributions of Deformable DETR models trained with and without TPRD across different distillation baselines: (a) DETRDistill, (b) D$^{3}$ETR, (c) QSKD, and (d) KD-DETR. Each shaded area represents the contribution of correcting a specific error type: Cls – classification error, Loc – localization error, Both – joint classification and localization error, Dupe – duplicate predictions, Bkg – false positives on background, and Miss – missed detections.}
    \label{fig:fig9}
\end{figure*}

\subsubsection{Analysis of training dynamics}
Fig.~\ref{fig:fig10} illustrates the training curves of different DETR distillation methods, both with and without applying TPRD. Across all methods, TPRD consistently accelerates convergence and yields higher final performance. In the early training phase, curves with TPRD rise more steadily, suggesting that refining the teacher’s supervision signals helps stabilize optimization and mitigate the influence of noisy gradients. In later epochs, methods with TPRD maintain a clear performance margin over their baselines, as highlighted in the zoomed-in subplot. This reflects the fact that  TPRD not only improves final mAP but also enhances the convergence behaviour and stability of DETR distillation.

\subsubsection{Distribution of different error types}
To better understand the source of performance gains brought by TPRD, we analyze different error types using TIDE~\cite{bolya2020tide} on Deformable DETR under various distillation methods. This analysis decomposes mAP into classification (Cls), localization (Loc), duplicate detections (Dupe), background false positives (Bkg), and missed detections (Miss).

As shown in Fig.~\ref{fig:fig9}, the shaded regions of the precision–recall curves indicate the mAP improvement obtained by correcting each error type, with values in parentheses reporting the corresponding gains. TPRD consistently reduces multiple error types across all methods, with the most pronounced improvements observed in classification errors, duplicate detections, and missed detections. For instance, under DETRDistill, Cls, Dupe, and Miss are reduced by 0.34\%, 0.13\%, and 0.69\%, respectively. Similar trends are observed for $\text{D}^3$ETR (0.33\%, 0.10\%, 0.36\%) and QSKD (0.15\%, 0.12\%, 0.68\%), while KD-DETR exhibits smaller yet consistent reductions across all error categories. These findings empirically validate our design motivation: by refining degraded positives and suppressing noisy negatives, TPRD delivers cleaner and more reliable supervision signals, thereby leading to more accurate and stable detection performance.

\begin{table}[t]
    \centering
    \footnotesize
    \caption{Training time (GPU hours) per epoch of different methods before and after applying TPRD, measured on A800 GPUs with a total batch size of 16.}
    \begin{tabular}{cccccc}
        \toprule
        Methods & DETRDistill & $\text{D}^3$ETR & QSKD & KD-DETR & Avg \\
        \midrule
        original & 1.67 & 1.83 & 1.83 & 1.04 & 1.59 \\
        +ours & 1.93 & 2.03 & 2.19 & 1.4 & 1.89 \\
        $\Delta$ & +0.26 & +0.20 & +0.36 & +0.36 & +0.30 \\
        \bottomrule
    \end{tabular}
    \label{tab:tab9}
\end{table}

\begin{figure}
    \centering
    \includegraphics[width=0.5\textwidth]{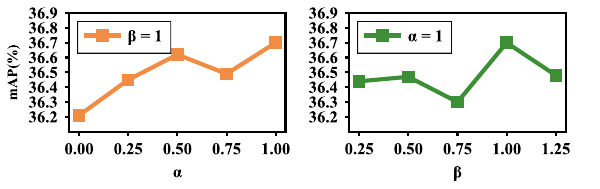}
    \caption{Sensitivity analysis of the hyperparameters $\alpha$ and $\beta$ on Deformable DETR. When varying $\beta$, $\alpha$ is fixed to 1; when varying $\alpha$, $\beta$ is fixed to 1. All results are obtained using a 12-epoch training schedule.}
    \label{fig:fig12}
\end{figure}

\subsubsection{Training time analysis} As a plug-and-play module, it is important to assess the additional training cost introduced by our method. Table~\ref{tab:tab9} presents the training time (GPU hours) per epoch for different distillation methods, both before and after applying TPRD. As shown, TPRD introduces only a modest increase in training time across various DETR distillation frameworks. On average, the training time per epoch increases by approximately 0.3 hours, less than a 20\% overhead relative to the respective baselines. This additional cost primarily arises from searching earlier decoder stages for better predictions. Notably, this computation is required only during training and can be entirely removed during inference, adding neither extra latency nor parameters. We consider this minor cost as a reasonable trade-off for the consistent accuracy improvements observed across different methods. In practice, the overhead can be further reduced by caching the best predictions for each stage in advance, without affecting model performance.

\subsubsection{Sensitivity analysis of hyperparameters $\alpha$ and $\beta$} We investigate the sensitivity of TPRD to the hyperparameters $\alpha$ and $\beta$, which control the relative contributions of target-class and non-target-class distillation in Eq.~(\ref{eq:13}). All other experimental settings are kept unchanged, and the student is trained using a 12-epoch schedule. As shown in Fig.~\ref{fig:fig12}, the performance remains relatively stable over the evaluated parameter ranges. When $\alpha$ varies from 0 to 1, the mAP ranges from 36.2 to 36.7, corresponding to a maximum variation of 0.5 AP. When $\beta$ varies from 0.25 to 1.25, the mAP ranges from 36.3 to 36.7, with a maximum variation of 0.4 AP. These results indicate that TPRD is not highly sensitive to the choice of $\alpha$ and $\beta$ within the investigated ranges. Based on the best observed performance and consistency with the default DKD setting, we set $\alpha=1$ and $\beta=1$ in all remaining experiments.

\begin{figure*}[t]
    \centering
    \includegraphics[width=\textwidth]{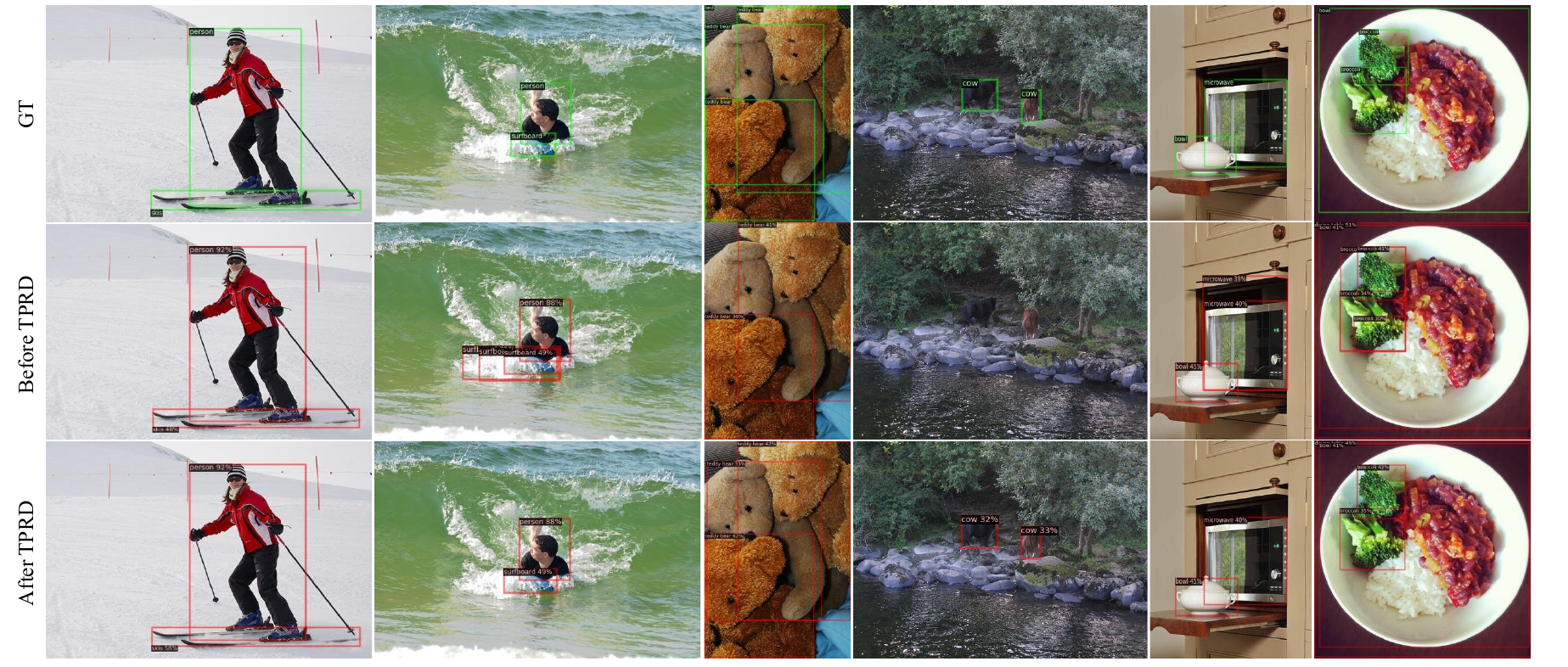}
    \caption{Qualitative comparison of the teacher model’s supervision signals before and after applying TPRD. From top to bottom: ground truth, teacher predictions before TPRD, and reconstructed predictions after TPRD. The improved results after TPRD indicate that our method effectively enhances the teacher’s supervision quality.}
    \label{fig:fig11}
\end{figure*}

\subsection{Visualization}
\subsubsection{Visualization of the cross-attention maps} 
\begin{figure}[t]
    \centering
    \includegraphics[width=0.5\textwidth]{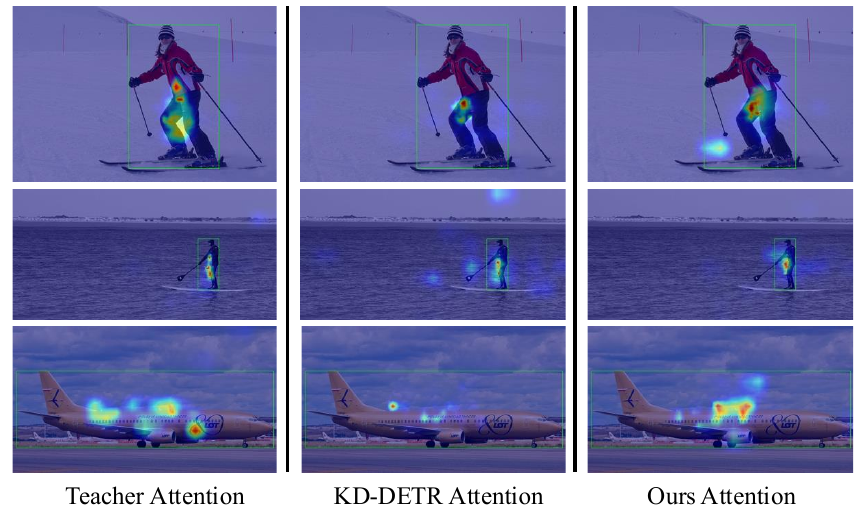}
    \caption{Visualization of the cross-attention maps from the last decoder stage of Deformable DETR. From left to right are results from the teacher model, KD-DETR, and our method. For each image, one positive prediction is randomly selected for visualization.}
    \label{fig:fig13}
\end{figure}

We visualize the cross-attention maps of the last decoder stage to verify that our method can better capture the teacher model’s knowledge. These maps explicitly indicate which image regions the model attends to, enabling intuitive comparisons of attention behavior between models. As shown in Fig.~\ref{fig:fig13}, KD-DETR often attends either too sparsely (e.g., the first and third rows) or too broadly (e.g., the second row). In contrast, our method exhibits attention patterns more consistent with the teacher model, providing qualitative evidence that TPRD enables more effective knowledge transfer.

\subsubsection{Visualization of the teacher model’s supervision signals before and after applying TPRD} In the previous section, we quantitatively demonstrated that our method improves the teacher model’s prediction quality. Here, we complement this analysis with qualitative results. As shown in Fig.~\ref{fig:fig11}, we visualize the teacher’s predictions before and after applying our method. Each row displays the ground truth, the teacher’s predictions before TPRD, and those after TPRD, respectively. In the first column, our method raises the confidence score of the skis from 0.48 to 0.58; in the second column, it eliminates several redundant detections of the surfboard; and in the fourth column, it successfully identifies two additional objects that were missed by the original teacher. The comparison reveals that TPRD substantially improves the accuracy of positive predictions while effectively suppressing noisy negative samples, yielding cleaner and more reliable supervision from the teacher model. Notice that this visualization illustrates how TPRD constructs cleaner positive and negative supervision signals for distillation, instead of demonstrating an improvement in teacher inference performance.

\section{Discussion}
\textbf{Limitations \& Future Work}. 
Despite its effectiveness, TPRD still has several limitations. Our experiments mainly focus on offline distillation, and TPRD introduces additional training cost due to the search over earlier decoder stages, although it adds no parameters or inference cost. Exploring more efficient target search strategies and extending TPRD to EMA-based or online distillation frameworks are promising directions for future work. 

\textbf{Difference from SQR}. SQR and TPRD both exploit intermediate decoder information, but they address different problems. SQR is a detector training strategy that recollects or replaces object queries to improve the optimization of the detector itself. Its primary objective is to provide better query representations for subsequent decoder processing. In contrast, TPRD is a teacher-side refinement mechanism designed specifically for knowledge distillation. It keeps the teacher parameters and decoder queries unchanged, and selectively reconstructs the teacher’s distillation targets by comparing predictions across decoder stages. Therefore, SQR modifies the query flow used for detector optimization, whereas TPRD modifies the supervision signal transferred from the teacher to the student. The two methods are conceptually complementary.

\section{Conclusion}
In this paper, we propose a plug-and-play module that consistently improves the performance of existing DETR distillation methods. Our study begins by revealing that, due to DETRs' cascading update mechanism, the prediction quality of DETR models fluctuates during training, leading to suboptimal supervision signals in previous methods. Based on this observation, we introduce Teacher Prediction Refinement Distillation (TPRD) to refine the teacher model’s supervision signals. Specifically, Positive Prediction Correction (PPC) corrects degraded positives to provide more accurate guidance, while Negative Prediction Suppression (NPS) suppresses overconfident negatives to avoid misleading supervision. In addition, we propose a Maximum Dark Knowledge Preservation (MDKP) module to retain the teacher’s inter-class relations throughout this refinement process. Extensive experiments on multiple DETR variants, various DETR distillation methods, and different teacher-student pairs verify the effectiveness and robustness ability of our approach. We hope this work offers a new perspective for advancing DETR distillation research.

\bibliographystyle{IEEEtran}
\bibliography{sample}










\newpage

\vfill

\end{document}